\documentclass[pmlr]{jmlr}

\RequirePackage{graphicx}
 \usepackage{booktabs}
 \usepackage{hyperref}

 \usepackage{multirow}
\usepackage{longtable}
 \usepackage{bm}

\makeatletter
\def\set@curr@file#1{\def\@curr@file{#1}} 
\makeatother
\usepackage{siunitx}

\theorembodyfont{\upshape}
\theoremheaderfont{\scshape}
\theorempostheader{:}
\theoremsep{\newline}

\usepackage{cleveref}
\usepackage{enumitem}

\title[SHIFT]{SHIFT: Survival Prediction from Incomplete and Heterogeneous Genomic Data}

\author{\Name{Muhammet Sami Yavuz}
       \Email{sami.yavuz@tum.de}\\ 
       \addr AI for Image-Guided Diagnosis and Therapy, School of Medicine and Health,\\
        Technical University of Munich (TUM), Munich, Germany\\
        Munich Center for Machine Learning (MCML)
        \AND
        \Name{Ayhan Can Erdur}
        \Email{can.erdur@tum.de}\\
        \addr Department of Radiation Oncology, TUM University Hospital\\
        \addr Chair for AI in Healthcare and Medicine,\\
        Technical University of Munich (TUM), Munich, Germany
        \AND
        \Name{Sabri Mustafa Kahya}
        \Email{mustafa.kahya@tum.de}\\
        \addr Technical University of Munich (TUM), Munich, Germany
        \AND
        \Name{Benedikt Wiestler}
        \Email{b.wiestler@tum.de}\\
       \addr AI for Image-Guided Diagnosis and Therapy, School of Medicine and Health,\\
        Technical University of Munich (TUM), Munich, Germany\\
        Munich Center for Machine Learning (MCML)
        \AND
        \Name{Jana Lipkova}
        \Email{jlipkova@hs.uci.edu}\\
        \addr Department of Pathology, School of Medicine,  \\
        Department of Biomedical Engineering, School of Engineering, \\
        University of California Irvine, Irvine, CA, USA}

\begin{document}

\maketitle

\begin{abstract}
Genomic prediction models often fail to transfer across institutions because sequencing panels differ across sites, creating structural feature missingness at deployment. Existing approaches to this challenge typically restrict analysis to genes shared across cohorts, exclude patients with incomplete profiles, or rely on test-time imputation, all of which can reduce robustness and limit the use of multi-center data. We propose \textit{Survival prediction Handling Incomplete Features using Transformer} (\textbf{SHIFT}), a missingness-aware survival model that directly predicts from incomplete genomic inputs without test-time imputation. SHIFT represents each genomic feature separately and uses masked self-attention, along with a feature-availability mask, so that predictions are based only on observed inputs. Further, we introduce variable-rate feature masking during training to improve robustness to heterogeneous missingness patterns. We evaluate the approach on glioblastoma and lung squamous cell carcinoma with external validation across multiple cohorts, including a challenging setting with severe cross-cohort panel mismatch. Across these settings, SHIFT shows strong generalization and compares favorably with standard survival baselines and imputation-based approaches, while using a single model across differing feature sets. We also find that incorporating patients from incomplete cohorts during development can improve performance on external data, suggesting that partially observed cohorts need not be excluded from model building. These results support missingness-aware modeling as a practical strategy for multi-center survival prediction in precision oncology.
\end{abstract}

\section{Introduction}
\begin{figure*}[t!]
\centering
\includegraphics[width=0.95\linewidth]{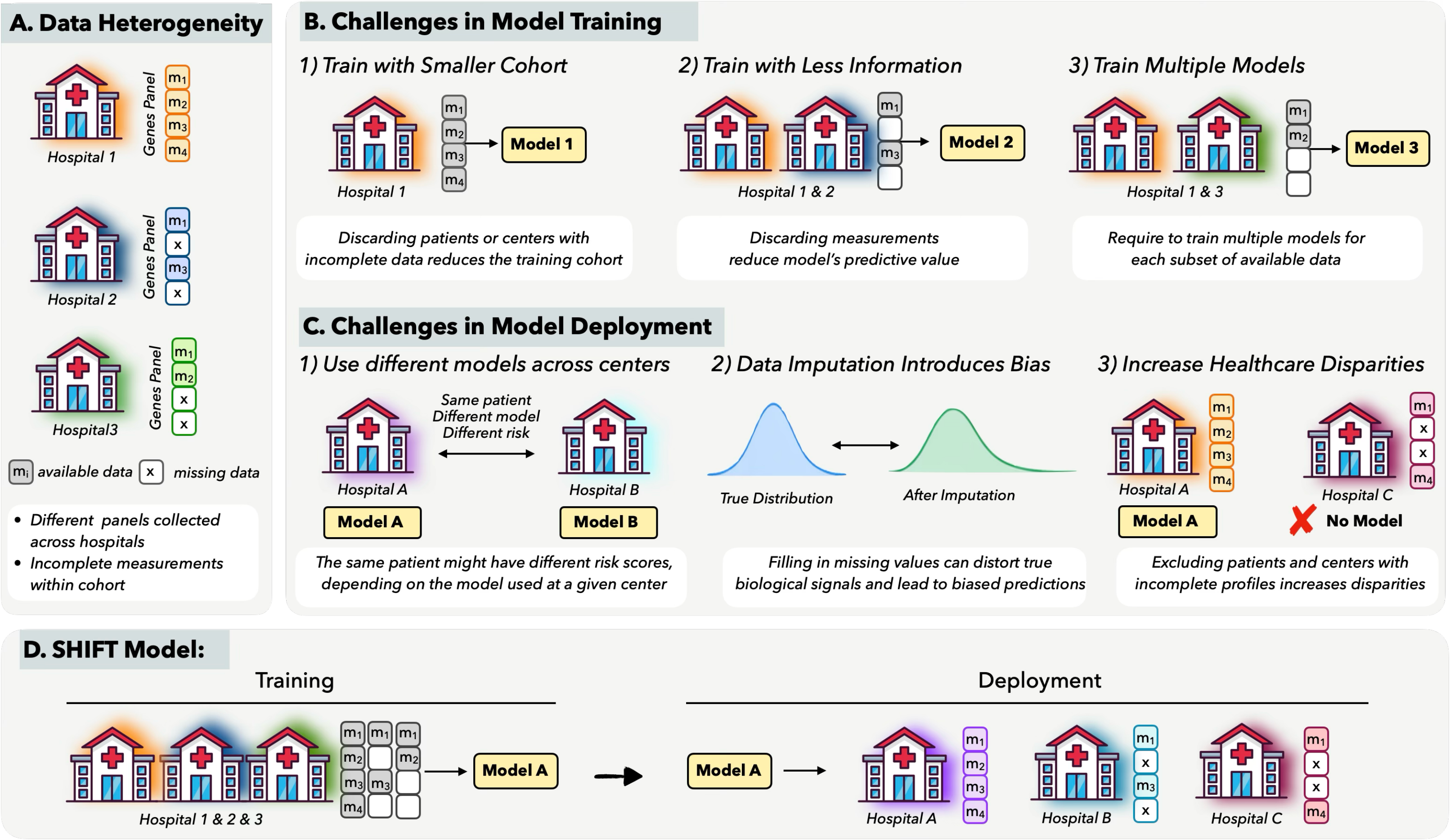}
\caption{\small \textbf{Addressing Challenges in Cross-Center Genomic Data Heterogeneity.} (A) Differences in data collection can result in substantial variability in the measured set of genes and incompatibilities across cohorts. (B–C) This heterogeneity introduces challenges for model training and deployment. (D) SHIFT addresses these challenges by enabling robust learning and inference from incomplete data using a single unified model.}
\label{fig:overview}
\end{figure*}

Genomic data play a central role in precision medicine by providing molecular insights that inform prognosis, risk stratification, and treatment planning. Recent advances in artificial intelligence (AI) and machine learning (ML) have demonstrated their feasibility for predicting patient survival from genomic profiles \citep{yousefi2017predicting,chaudhary2018deep,lee2023deep,wiegrebe2024deep}. However, the development and deployment of such models remain challenging due to substantial heterogeneity in genomic data across centers. Differences in sequencing platforms, diagnostic workflows, and resource availability result in substantial variability in the set of measured genes, ranging from broad sequencing assays to targeted panels that measure only a subset of genes \citep{chen2024unlocking,flores2023missing,garcia2017validation}. This variability in feature availability poses significant challenges for AI model development and deployment, as illustrated in \Cref{fig:overview}.

Most existing AI models assume fully observed inputs and cannot robustly handle missing data. As a result, patients with incomplete profiles are often excluded from studies, reducing the size and diversity of training cohorts and potentially amplifying selection bias \citep{lipkova2022artificial}. Alternatively, models may be trained only on the subset of features shared across cohorts, but this strategy might discard relevant genomics information and limit model performance. Data imputation is another commonly used strategy; however, it can distort underlying biological signals and introduce bias into model predictions \citep{donders2006gentle,alwateer2024missing}. This is further exacerbated by the fact that missingness in genomic data is often structural rather than random, as different institutions measure different sets of genes. In such settings, a large proportion of features may need to be imputed, further amplifying these issues. In practice, this often necessitates training separate models for each subset of available genes, which is difficult to scale across centers. The deployment of different models across centers may further exacerbate the disparities in healthcare \citep{chen2023algorithmic}.

To address these challenges, we propose \textit{Survival prediction Handling Incomplete Features using Transformer} (\textbf{SHIFT}), a transformer-based model for survival prediction from incomplete tabular genomic data. SHIFT uses a masked self-attention mechanism that dynamically shifts the model’s focus to the genomics information available for each patient, while handling missing values. SHIFT is trained with randomly varying rates of feature missingness, encouraging the model to learn inter-feature relationships from partially observed inputs and to leverage these dependencies when encountering naturally incomplete profiles at inference time. This design enables SHIFT to be trained and deployed across heterogeneous genomic panels without site-specific retraining or imputation. We evaluate the proposed framework across two cancer types, glioblastoma (GBM) and lung squamous cell carcinoma (LUSC), assessing its predictive performance and generalization to external cohorts. Our evaluation considers cohorts with substantial structural missingness arising from cross-center data heterogeneity, reflecting real-world clinical settings. Across these scenarios, we demonstrate that SHIFT can serve as a practical alternative to traditional imputation-based approaches. We further show that even highly incomplete cohorts can provide informative signals during training, improving overall model performance. Our main contributions can be summarized as follows:
\begin{enumerate}[leftmargin=*, noitemsep, topsep=3pt]
\item  We introduce SHIFT, a transformer-based model for survival prediction from incomplete genomics data without imputation. 
\item We leverage a variable-rate masking (VRM) strategy during training that improves robustness to cross-cohort shift.
\item We show that severely incomplete cohorts can still contribute useful signal during model development, supporting more inclusive multi-center modeling.

\end{enumerate}

\subsection*{Generalizable Insights about Machine Learning in the Context of Healthcare}

Our study yields three broader insights for machine learning in healthcare: First, a single model can be used for survival prediction across institutions despite the variability in the available data. This is important in healthcare settings, where missingness is often structural and arises from differences in assays and clinical workflows across institutions. Second, model training with simulated missing data can improve robustness and predictive performance. Our experiments show that training with our VRM strategy not only improves performance under incomplete data at inference time, but also acts as a form of data augmentation and regularization, enhancing performance even on cohorts without missing data. Third, incomplete cohorts should not automatically be excluded from model development. Even cohorts with substantial feature absence can provide useful signals when missingness is handled directly by the model. These findings suggest that healthcare ML systems may become more robust and deployable when designed to learn from heterogeneous features for each patient rather than assuming fully standardized inputs.

\section{Related Work}
\textbf{Survival prediction models.}
Over the years, a plethora of models for survival prediction have been developed. Arguably, the Cox proportional hazards model \citep{cox1972regression} remains the
standard clinical benchmark, assuming a log-linear hazard ratio, and is widely used in many clinical studies.
Modern approaches typically leverage data-driven learning instead of parametrized models and re-purpose them for survival analysis:
Random Survival Forests (RSF) \citep{Ishwaran_2008} extend random forests to right-censored data via log-rank splitting, capturing non-linear
effects without distributional assumptions.
DeepSurv \citep{katzman2018deepsurv} replaces the linear risk function with a deep neural network that optimizes the Cox partial likelihood.
DeepHit \citep{lee2018deephit} reframes survival as discrete-time multi-label classification over a set of predefined time intervals, avoiding the proportional hazards assumption.
Self-normalizing networks (SNNs) with SELU activations \citep{klambauer2017self} are well-suited to tabular genomic data \citep{chen2022pan,chen2021multimodal,jaume2024modeling} because they allow training more stable models without batch normalization and represent the current state-of-the-art approach. A detailed review of ML models for survival prediction in oncology is provided in \citep{wiegrebe2024deep}.

\textbf{Handling missing data for incomplete genomic inputs.} Methods to address missing data generally fall into two categories: imputation and adaptation.

\textbf{1) Imputation methods} approximate or fill in missing data during preprocessing. Common approaches include k-Nearest Neighbor (KNN) \citep{beretta2016nearest, pujianto2019k}, mean imputation \citep{petrazzini2021evaluation}, and imputation via regression trees \citep{burgette2010multiple}. However, these methods are most appropriate when missingness is relatively sparse and approximately missing-at-random. In cross-cohort genomics, missingness is often structural, and large feature blocks may be absent due to differences in sequencing panels \citep{flores2023missing, chen2024unlocking}. Synthetic data generation has emerged as an alternative strategy \citep{chen2021synthetic,yi2019generative,forestier2017generating}, particularly for imputing images and time-series data. However, its application to prognostic genomic tasks remains problematic, as there is no guarantee that synthetic data preserves clinically relevant signals, which are more complex than the mere presence or absence of specific mutations. These methods are also prone to hallucination \citep{cohen2018distribution}, compromising the reliability of downstream survival predictions. Generally, imputation-based pipelines often require choosing and fitting a separate strategy for each deployment setting, limiting their practicality under highly heterogeneous genomic profiles.

\textbf{2) Adaptation strategies} modify model architectures to natively accommodate incomplete inputs. A promising example is transformer-based attention masking. In masked self-attention, tokens corresponding to missing features are excluded from attention interactions, so the model constructs representations from the observed subset alone. This idea has shown promise in time-series settings with missing values \citep{fan2024multiscale, neog2025masking, zhang2024improved} and in tabular classification \citep{caruso2024not}. The closest prior work to ours is \citep{caruso2024not}, which applies masked attention to tabular data. However, prior work focuses on simple binary classification tasks, where a small subset of input variables is often sufficient for correct predictions. These works further consider only random data absence in a small subset of inputs. Methods for handling missing data are reviewed in detail \citep{heymans2022handling,zhou2024review,joel2022review}, and it remains an open challenge in oncology \citep{lipkova2022artificial,flores2023missing}.

The SHIFT model extends the prior works to enable survival prediction from incomplete genomic data, addressing the real-world challenge of structural missingness across institutions.
Our work focuses heavily on model generalization to external cohorts and studies whether variable data absence during training improves generalization under heterogeneous deployment conditions.

\section{Methods}

\subsection{Problem Setup and Discrete-Time Survival Prediction Objective}

Our goal is to predict patient survival from genomic features
$\mathbf{x}^{(i)} \in \mathbb{R}^{d}$, where $d$ denotes the number of
genomic variables and $i \in \{1,\dots,P\}$ indexes patients.
Following standard survival-analysis notation \citep{zadeh2020bias},
each patient is associated with: (1) an observed survival time
$T^{(i)}_{\mathrm{cont}}$, defined as the time from diagnosis to either
death or last follow-up; and (2) a censoring indicator
$c^{(i)} \in \{0,1\}$.

Following prior deep survival models
\citep{vale2021long,chen2021multimodal,jaume2024modeling}, we cast
survival prediction as a discrete-time problem rather than directly
regressing the continuous survival time.
Specifically, for each training fold, we partition the continuous time
axis into $K=4$ non-overlapping intervals
$[t_0,t_1), [t_1,t_2), [t_2,t_3), [t_3,t_4)$
using the quartiles of uncensored survival times in the corresponding
training split. Similar to the prior study \citep{chen2020pathomic}, we use quartile-based binning (K = 4), which provides sufficient event count per interval despite smaller cohort sizes. We then define the discrete survival interval
$Y^{(i)} \in \{0,1,2,3\}$ such that
\[
Y^{(i)} = j
\quad \text{if} \quad
T^{(i)}_{\mathrm{cont}} \in [t_j, t_{j+1}),
\qquad j \in \{0,1,2,3\}.
\]

\noindent Let $h_j(\mathbf{x}^{(i)})$ denote the discrete hazard for interval $j$.
Given model logits $\hat{y}^{(i)}_j$, we define
\[
h_j(\mathbf{x}^{(i)}) = \sigma\!\left(\hat{y}^{(i)}_j\right),
\]

\noindent where $\sigma(\cdot)$ is the sigmoid function. Intuitively, $h_j(\mathbf{x}^{(i)})$ is the probability that patient $i$ experiences the event during interval $[t_j,t_{j+1})$. The corresponding discrete survival function~is
\[
S_j(\mathbf{x}^{(i)}) =
\prod_{k=0}^{j} \left(1 - h_k(\mathbf{x}^{(i)})\right),
\qquad
S_{-1}(\mathbf{x}^{(i)}) \triangleq 1,
\]

\noindent which represents the probability of surviving through interval $j$. We optimize the discrete-time negative log-likelihood loss
\citep{zadeh2020bias}, which can be written as three terms:
\begin{align}
\mathcal{L}^{(i)}_1
&=
- c^{(i)} \log S_{Y^{(i)}}(\mathbf{x}^{(i)}),
\label{eq:l1} \\
\mathcal{L}^{(i)}_2
&=
- (1-c^{(i)}) \log S_{Y^{(i)}-1}(\mathbf{x}^{(i)}),
\label{eq:l2} \\
\mathcal{L}^{(i)}_3
&=
- (1-c^{(i)}) \log h_{Y^{(i)}}(\mathbf{x}^{(i)}).
\label{eq:l3}
\end{align}
Here, $\mathcal{L}^{(i)}_1$ encourages high survival probability up to the censoring time for censored patients, while $\mathcal{L}^{(i)}_2$ and
$\mathcal{L}^{(i)}_3$ encourages uncensored patients to survive until the
interval preceding the event and to place high hazard on the event
interval itself. Following \citep{zadeh2020bias}, we downweight the censored term using
a scalar $\beta < 1$. In all experiments, we set $\beta=0.6$, yielding the overall objective

\begin{equation}
\begin{aligned}
\mathcal{L}
=
\frac{1}{P}
\sum_{i=1}^{P}
\left[
(1-\beta)\mathcal{L}^{(i)}_1
+
\mathcal{L}^{(i)}_2
+
\mathcal{L}^{(i)}_3
\right].
\end{aligned}
\label{equation}
\
\end{equation}
This assigns a weight of $0.4$ to the censored term and a weight of $1.0$ to each uncensored term. For risk ranking, we define the patient-level risk score as
\[
r(\mathbf{x}^{(i)}) = - \sum_{j=0}^{K-1} S_j(\mathbf{x}^{(i)}),
\]
so that patients with lower predicted survival receive higher risk
scores, enabling patient ranking for C-index computation.

\begin{figure*}[t!]
\centering
\includegraphics[width=0.95\linewidth]{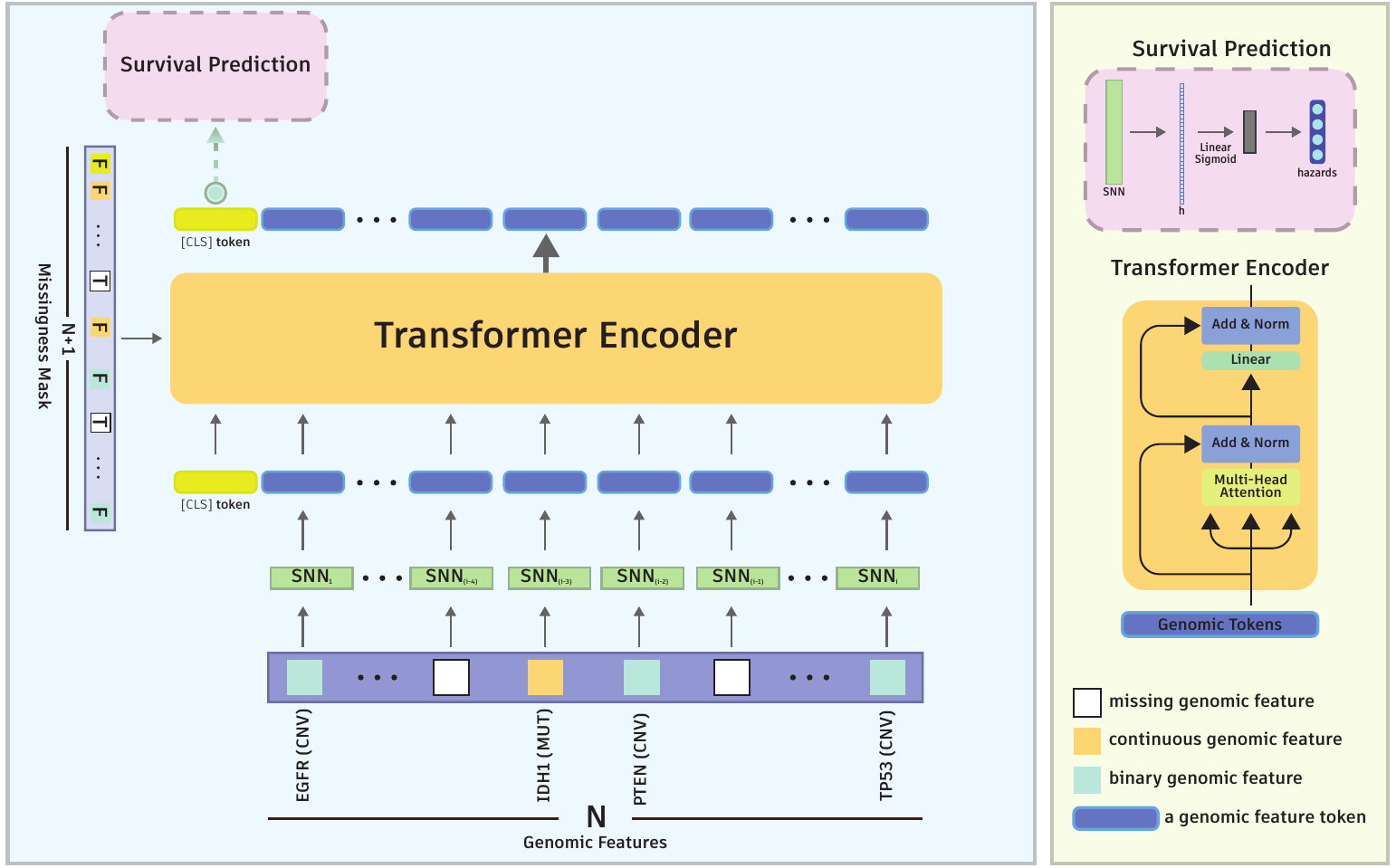}
\caption{\small \textbf{Overview of the SHIFT model.}
SHIFT consists of three main components (described in detail in Sec.~\ref{sec:shift-arch}):
\textit{(i) feature embedding block} which maps each genomic variable to a token representation, \textit{(ii) masked transformer encoder} that leverages the missingness mask to focus attention on available genomic information and to learn prognostically relevant representations from available inputs, and \textit{(iii) survival prediction block}, which predicts patient survival outcomes from the [CLS] token.}
\vspace{-4mm}
\label{fig:shift-main}
\end{figure*}

\subsection{SHIFT: Missingness-Aware Transformer for Survival Prediction}
\label{sec:shift-arch}
SHIFT is a transformer-based model for survival prediction from incomplete genomic data. It consists of three main parts: \textit{(i) feature embedding block}, which maps each genomic variable to a token representation, \textit{(ii) masked transformer encoder} that shifts the model's attention to available genomics features for each patient, and \textit{(iii) survival prediction block}, which predicts risk scores for each patient. The model architecture is shown in \cref{fig:shift-main} and described below:

\paragraph{(i) Feature embedding block:} Instead of processing the full genomic vector jointly, SHIFT maps each input genomic variable independently into a token representation. For this mapping, we use the SNN block defined as:
\[
\mathrm{SNN}(n_{\mathrm{in}}, n_{\mathrm{out}})
=
\mathrm{Linear}(n_{\mathrm{in}} \rightarrow n_{\mathrm{out}})
\rightarrow \mathrm{SELU}
\rightarrow \mathrm{AlphaDropout}, 
\]
where $n_{\mathrm{in}}$ and $n_{\mathrm{out}}$ are the input and output dimensions of the block. Thus, SHIFT embeds each genomic variable feature $x_l \in \mathbb{R}$ independently using a dedicated $\mathrm{SNN}(1, D)$, where $D=128$ and
$l \in \{1,\dots,d\}$ indexes genomic features. This produces a sequence of feature tokens:
\[
\mathbf{E} = [\mathbf{e}_1,\dots,\mathbf{e}_d] \in \mathbb{R}^{d \times D}.
\]
Because each feature has its own embedding parameters, SHIFT can learn feature-specific representations while naturally accommodating a different set of observed features at inference time.
\\
Missing features are zero-filled before embedding, and marked by a binary missingness mask $\mathbf{M} \in \{\mathrm{True},\mathrm{False}\}^{d+1}$, where $M_l=\mathrm{True}$ indicates that feature $l$ is missing. Missing features are zero-filled only as placeholders. Since their tokens are masked in the attention mechanism, they do not contribute to context aggregation. A learnable class token
$[\mathrm{CLS}] \in \mathbb{R}^{D}$ is prepended to the token sequence, with a corresponding mask entry fixed to $M_0=\mathrm{False}$.

\paragraph{(ii) Masked transformer encoder.}
The token sequence is processed by a 2-layer Transformer encoder
\citep{vaswani2017attention} with hidden dimension $D=128$ and $H=4$ attention heads. We use the missingness mask $\mathbf{M}$ as a key-padding mask so that tokens corresponding to missing features are excluded from attention. As a result, the $[\mathrm{CLS}]$ token aggregates information only from observed genomic variables. We omit positional embeddings because genomic feature identity is already captured by the feature-specific embedding parameters, and the feature set has no
natural sequential order.

\paragraph{(iii) Survival prediction block.}
The final $[\mathrm{CLS}]$ representation is passed through an
$\mathrm{SNN}(128,256)$, followed by a linear layer
$(256 \rightarrow K)$ to produce interval-specific logits. A sigmoid activation converts these logits into discrete hazard probabilities for the $K$ survival intervals. The full model is trained end-to-end using the
discrete-time negative log-likelihood objective in Eq. 4.


\subsection{Training with Variable-Rate Feature Masking}
Beyond the base SHIFT model, we leverage variable-rate masking (VRM), a training-time masking strategy that enhances model robustness to missing data. For each epoch during the model training, we randomly sample a fraction of inputs to be masked for each patient $i$, given as:
\[
k_i \sim \mathrm{Uniform}\{0,1,\dots,\lfloor f \cdot d \rfloor\},
\]
where $d$ is the number of genomic features and $f \in [0,1]$ controls
the maximum masking fraction.
We then uniformly sample $k_i$ feature indices without replacement and mark them as missing.
Thus, each training sample is exposed to a different masking level, ranging from fully observed inputs to at most $(100\times f)\%$ of the features masked. The value $f=0.0$ implies that the VRM approach is disabled (termed `SHIFT w/o VRM'), whereas $f=1.0$ indicates that all features except the class token $[\mathrm{CLS}]$ may be masked. We use $f=0.5$ as the default setting (termed `SHIFT-VRM') because it provides the most consistent performance across cohorts in the masking-fraction ablation (Sec.~\ref{sec:ablation-f} for an extensive ablation).
VRM is intended to expose the model to diverse observed-feature subsets during training, making the model more robust to the variability in available data. The VRM strategy is only used during the model training. At inference time, naturally missing features are handled natively through the missingness mask, without any test-time imputation.

\subsection{Baseline Methods}
\label{sec:baseline-methods}
We compare SHIFT against six baseline methods spanning traditional and deep survival modeling:
XGBoost with a Cox survival objective \citep{10.1145/2939672.2939785},
CoxPH \citep{cox1972regression},
Random Survival Forest (RSF) \citep{Ishwaran_2008},
DeepSurv \citep{katzman2018deepsurv},
DeepHit \citep{lee2018deephit},
and an SNN baseline model \citep{klambauer2017self,chen2022pan}. For CoxPH, RSF, XGBoost-Cox, DeepSurv, and DeepHit models, we use their standard implementation. For the SNN baseline, we used the implementation described in \citep{chen2022pan}. Accordingly, the model jointly processes the full genomic feature vector, without tokenization or attention.
Its backbone consists of three consecutive SNN blocks with hidden dimension 256 ($d \rightarrow 256 \rightarrow 256 \rightarrow 256$).
The resulting representation is passed to a two-layer survival head:
$\mathrm{Linear}(256 \rightarrow 128)$ followed by ReLU, Dropout(0.7),
and $\mathrm{Linear}(128 \rightarrow K)$.
A sigmoid activation is applied to the final logits to obtain discrete
hazard probabilities over the $K$ survival intervals. 

Since baseline models require complete feature vectors at inference time, we integrate test-time imputation to handle missing features. For this purpose, we employ two common strategies: KNN and mean imputation. Under KNN imputation, missing values are estimated using the $k=5$ nearest neighbors fit on the training data. With mean imputation, each missing feature is replaced by its feature-wise mean estimated from the corresponding training data. XGBoost handles missing values naturally and therefore does not require imputation.

\section{Cohorts}
\subsection{Cohort Selection}
We evaluate SHIFT on two cancers, GBM and LUSC, using public and private cohorts.

GBM includes three cohorts: 1) The Cancer
Genome Atlas (TCGA) \citep{TCGA} cohort with 420 patients,  2) Clinical Proteomic Tumor Analysis Consortium (CPTAC)
\citep{edwards2015cptac} cohort with 93 patients, and 3) a private cohort of 138 patients sourced from Germany. Survival information for all patients is reported in months. All cohorts for GBM  share the same set of 36 genomic variables: continuous copy-number variation (CNV) measurements for 34 genes relevant to GBM biology, including \textit{EGFR}, \textit{PTEN}, \textit{PDGFRA}, \textit{CDK4}, \textit{MDM2}, \textit{RB1}, \textit{TP53}, and \textit{NF1}, together with binary indicators for \textit{IDH1} mutation and 1p/19q codeletion. There is no structural data incompleteness in these datasets; they are used to test the model's robustness and to evaluate the benefits of the VRM strategy.

LUSC includes data from three cohorts: 1) the TCGA cohort, including 470 patients, 2) the CPTAC, with 101 patients, and 3) a private cohort of 102 patients from the US. Survival information for all patients is reported in months. The LUSC molecular information consists of lung cancer-relevant mutations, including \textit{EGFR}, \textit{TP53}, \textit{KRAS}, and \textit{KEAP1}. However, the TCGA and CPTAC cohorts include 197 mutations, whereas the US cohort covers only 22 mutations available in TCGA and CPTAC. As a result, 175 of 197 features (88.8\%) are structurally absent in the US cohort. This setting reflects a real-world scenario in which different centers utilize significantly different sequencing panels.

\subsection{Data Preprocessing}
For GBM, CNV values are computed as
$\log_2(\text{observed}/\text{expected})$, where the expected copy number
corresponds to a diploid genome; positive values indicate amplification and negative values indicate deletion.
Mutation features are binary ($1$ for mutated, $0$ for wild-type). We apply z-score normalization to all genomic features, using the mean and standard deviation computed from the training data.

\subsection{Training Protocol} In our experiments, the TCGA cohort is used for model development, with stratified 5-fold cross-validation at an 80/20 train-validation split. Stratification is performed to preserve censoring proportions across folds. The external cohorts are used for evaluation or as an additional training cohort (only the US). For methods that require imputation, the imputer is fit on the training split of each fold only.

\textbf{Hyperparameters.}
We trained SHIFT with a batch size of 16, Adam
optimizer, learning rate $5 \times 10^{-4}$, and early stopping with
patience of 15 epochs based on validation C-index, for a maximum of 200 epochs.

\subsection{Evaluation Protocol}

We evaluate survival ranking performance using the concordance index (C-index), which measures the rank correlation between predicted risk scores and observed survival times. A higher C-index indicates better risk ranking, while $0.5$ corresponds to random predictions. We report the \textbf{mean performance} across 5 cross-validation folds, along with ensemble results that aggregate predictions from fold-specific models. In particular, \textbf{Ens-All} averages each patient’s predicted risk scores across all five models before computing a single cohort-level C-index. We also consider a \textbf{Top-3} ensemble, which averages predictions from the three best-performing models, ranked by validation C-index, before computing a cohort-level C-index. Model selection is based solely on validation performance within each fold, and external cohorts are never used for selection.

\section{Results}
To assess SHIFT's performance, we conducted three experiments evaluating: (1) non-inferiority to baseline models on complete datasets, (2) feasibility of model inference on cohorts with incomplete data, and (3) model training across centers with heterogeneous genomic profiles, reflecting real-world patterns of data missingness.

\subsection{SHIFT is Competitive to Baseline Models on Complete Datasets}

Tables~\ref{tab:gbm} and~\ref{tab:lusc-main} summarize performance in settings where training and external evaluation are carried out on aligned feature spaces: GBM is evaluated on CPTAC and the German cohorts using the 36 features, and LUSC is evaluated on CPTAC in the 197-feature setting. This serves as a control setting to test whether masking improves external robustness beyond handling literal feature absence.  Performance of all models is comparable to that reported in prior studies \cite{chen2022pan}.

On both GBM and LUSC, no single baseline consistently performs best. For example, while RSF achieves the best performance on LUSC-CPTAC, SNN outperforms it on GBM-CPTAC. Along this line, strong validation set performance does not necessarily translate into strong generalizability across test cohorts, indicating a nontrivial cross-cohort shift even when all features are observed at test time. In this setting, where all features are present in the test cohort, SHIFT w/o VRM already performs robustly, often beating the established baselines. However, its external transfer to the GBM-German  cohort is substantially weaker, indicating that architecture alone is not sufficient for robust cross-cohort generalization. Adding VRM, however, substantially improves cross-institutional generalization for both GBM and LUSC, yielding the highest C-index in all cohorts (usually Ens-All). Together, these results show that the SHIFT architecture is already competitive on full-feature data, and that variable-rate masking provides an additional gain in external robustness.

\begin{table}[t]
\centering
\caption{\textbf{Performance on GBM without missing data.} We report the C-index on the TCGA validation set and two external cohorts (CPTAC and German), with all models trained on the complete set of 36 features. Results are presented as mean $\pm$ standard deviation over 5 cross-validation folds. Top-3 denotes an ensemble of the three best-performing folds selected based on validation C-index, while Ens-All denotes an ensemble across all five fold-specific models. Bold indicates
the best result in each column.}
\label{tab:gbm}
\resizebox{\linewidth}{!}{
\begin{tabular}{lc|ccc|ccc}
\toprule
\textbf{Method} & \textbf{TCGA-Val} & \multicolumn{3}{c}{\textbf{CPTAC}} & \multicolumn{3}{c}{\textbf{German}} \\
\cmidrule(lr){3-5}\cmidrule(lr){6-8}
& Mean & Mean & Ens-All & Top-3 & Mean & Ens-All & Top-3 \\
\midrule
\multicolumn{8}{l}{\textit{Baselines}} \\
\midrule
XGBoost         & $0.5324_{\pm 0.02}$ & $0.5238_{\pm 0.04}$ & 0.4810 & 0.5594 & $0.5295_{\pm 0.03}$ & 0.5528 & 0.5310 \\
CoxPH           & $0.5115_{\pm 0.04}$ & $0.5157_{\pm 0.03}$ & 0.4795 & 0.5303 & $0.5308_{\pm 0.03}$ & 0.5192 & \textbf{0.5523} \\
RSF             & $0.5450_{\pm 0.03}$ & $0.5108_{\pm 0.02}$ & 0.4882 & 0.5102 & $0.5290_{\pm 0.02}$ & 0.5342 & 0.5326 \\
DeepSurv        & $0.5255_{\pm 0.02}$ & $0.4955_{\pm 0.03}$ & 0.4982 & 0.5024 & $\bm{0.5556_{\pm 0.03}}$ & 0.5567 & 0.5351 \\
DeepHit         & $0.4674_{\pm 0.05}$ & $0.4886_{\pm 0.03}$ & 0.4988 & 0.4926 & $0.4819_{\pm 0.03}$ & 0.4715 & 0.4818 \\
SNN             & $0.4975_{\pm 0.03}$ & $0.5342_{\pm 0.04}$ & 0.5347 & 0.5166 & $0.5222_{\pm 0.04}$ & 0.5422 & 0.5313 \\
\midrule
SHIFT w/o VRM   & $\bm{0.5493_{\pm 0.04}}$ & $0.5390_{\pm 0.02}$ & 0.5483 & 0.5441 & $0.4926_{\pm 0.04}$  & 0.5007 & 0.4755 \\
SHIFT-VRM       & $0.5327_{\pm 0.02}$ & $\bm{0.5530_{\pm0.01}}$ & \textbf{0.5827} & \textbf{0.5839} & $0.5309_{\pm0.01}$  & \textbf{0.5673} & 0.5470 \\
\bottomrule
\end{tabular}}
\end{table}

\begin{table}[t]
\centering
\caption{\textbf{Performance on LUSC without missing data.} We report the C-index on the TCGA validation set and CPTAC external cohort, all with the full set of 197 features. The C-index is presented as a mean $\pm$ std across all 5 folds, and as ensemble performance across all (Ens-All) and the three best (Top-3) folds. Bold indicates the best result in each column.}
\label{tab:lusc-main}

\begin{tabular}{lc|ccc}
\toprule
\textbf{Method} & \textbf{TCGA-Val} & \multicolumn{3}{c}{\textbf{CPTAC}} \\
\cmidrule(lr){3-5}
& Mean & Mean & Ens-All & Top-3 \\
\midrule

XGBoost       & $0.4673_{\pm 0.05}$ & $0.5521_{\pm 0.06}$ & 0.5490 & 0.5635 \\
CoxPH         & $0.4710_{\pm 0.06}$ & $0.5080_{\pm 0.02}$ & 0.5225 & 0.5172 \\
RSF           & $0.4676_{\pm 0.07}$ & $\bm{0.5631_{\pm 0.03}}$ & 0.5817 & \textbf{0.5795} \\
DeepSurv      & $0.4530_{\pm 0.04}$ & $0.5561_{\pm 0.01}$ & 0.5659 & 0.5542 \\
DeepHit       & $0.4935_{\pm 0.05}$ & $0.4819_{\pm 0.02}$ & 0.5097 & 0.4741 \\
SNN           & $\bm{0.5272_{\pm 0.03}}$ & $0.5399_{\pm 0.03}$ & 0.5368 & 0.5220 \\
\midrule
SHIFT w/o VRM & $0.4883_{\pm 0.02}$ & $0.5597_{\pm 0.03}$ & 0.5965 & 0.5598 \\
SHIFT-VRM     & $0.5170_{\pm 0.06}$ & $0.5600_{\pm 0.03}$ & \textbf{0.6021} & 0.5674 \\

\bottomrule
\end{tabular}

\end{table}
\begin{table}[t]
\centering

\caption{\textbf{Model performance on incomplete LUSC data.} All models are trained on the TCGA cohort with the full set of 197 features and tested on an external US cohort with only a subset of 22 features. Baseline models use KNN and  mean imputations (see Sec.~\ref{sec:baseline-methods}) to allow inference in the US cohort, while SHIFT models are evaluated without any imputation. 
}
\label{tab:lusc-us-full}
\resizebox{\linewidth}{!}{
\begin{tabular}{lc|ccc|ccc}
\toprule
\textbf{Method} & \textbf{TCGA-Val} &
\multicolumn{3}{c|}{\textbf{US (KNN)}} &
\multicolumn{3}{c}{\textbf{US (Mean)}} \\
\cmidrule(lr){3-5}\cmidrule(lr){6-8}
& Mean & Mean & Ens-All & Top-3 & Mean & Ens-All & Top-3 \\
\midrule
\multicolumn{8}{l}{\textit{Baselines}} \\
\midrule
XGBoost  & $0.4673_{\pm 0.05}$ & $\bm{0.5014_{\pm 0.02}}$ & \textbf{0.5195} & \textbf{0.4949} & $0.5014_{\pm 0.02}$ & 0.5195 & 0.4949 \\
CoxPH    & $0.4710_{\pm 0.06}$ & $0.4942_{\pm 0.03}$ & 0.4955 & 0.4818 & $0.5451_{\pm 0.01}$ & \textbf{0.5695} & 0.5404 \\
RSF      & $0.4676_{\pm 0.07}$ & $0.4702_{\pm 0.01}$ & 0.4369 & 0.4695 & $0.5541_{\pm 0.01}$ & 0.5530 & \textbf{0.5610} \\
DeepSurv & $0.4530_{\pm 0.04}$ & $0.4471_{\pm 0.01}$ & 0.4928 & 0.4436 & $\bm{0.5581_{\pm 0.01}}$ & 0.5514 & 0.5608 \\
DeepHit  & $0.4935_{\pm 0.05}$ & $0.4613_{\pm 0.01}$ & 0.4412 & 0.4624 & $0.4548_{\pm 0.01}$ & 0.5109 & 0.4531 \\
SNN      & $\bm{0.5272_{\pm 0.03}}$ & $0.4343_{\pm 0.02}$ & 0.4216 & 0.4231 & $0.4457_{\pm 0.02}$ & 0.4338 & 0.4360 \\
\midrule

\multicolumn{2}{c}{} &
\multicolumn{3}{c}{\textbf{US (No Imputation)}} &
\multicolumn{3}{c}{} \\
\cmidrule(lr){3-5}
& & Mean & Ens-All & Top-3 & \multicolumn{3}{c}{} \\
\midrule
SHIFT w/o VRM & $0.4883_{\pm 0.02}$ & $0.5102_{\pm 0.04}$ & 0.5548 & \textbf{0.5737} & \multicolumn{3}{c}{} \\
SHIFT-VRM     & $0.5170_{\pm 0.06}$ & $\bm{0.5370_{\pm 0.03}}$ & \textbf{0.5697} & 0.5712 & \multicolumn{3}{c}{} \\
\bottomrule
\end{tabular}}
\end{table}

\begin{table}[t]
\centering
\caption{\textbf{Model performance using the shared 22 LUSC features.} All methods are
trained and evaluated using only the 22 genomic features present in both
TCGA and US, with no imputation, except SHIFT-VRM (which is trained on all 197 features and can be natively applied). The C-index is presented as a mean $\pm$ std across all 5 folds, and as ensemble performance across all (Ens-All) and the three best (Top-3) folds. Bold
indicates the best result in each column.}
\label{tab:lusc-us-22}

\begin{tabular}{lc|ccc}
\toprule
\textbf{Method} & \textbf{TCGA-Val} &
\multicolumn{3}{c}{\textbf{US (22 features)}} \\
\cmidrule(lr){3-5}
& Mean & Mean & Ens-All & Top-3  \\
\midrule

XGBoost         & $0.4473_{\pm 0.04}$ & $0.4995_{\pm 0.02}$ & 0.4870 & 0.4873 \\
CoxPH           & $0.4638_{\pm 0.05}$ & $0.4867_{\pm 0.02}$ & 0.4968 & 0.5005 \\
RSF             & $0.4825_{\pm 0.04}$ & $0.5130_{\pm 0.01}$ & 0.5086 & 0.5012 \\
DeepSurv        & $0.4527_{\pm 0.04}$ & $0.4997_{\pm 0.02}$ & 0.4870 & 0.4943 \\
DeepHit         & $0.4908_{\pm 0.07}$ & $0.4887_{\pm 0.01}$ & 0.4612 & 0.4526 \\
SNN             & $\bm{0.5071_{\pm 0.03}}$ & $0.5148_{\pm 0.03}$ & 0.5281 & 0.5100 \\

\midrule
SHIFT w/o VRM   & $0.5026_{\pm 0.06}$ & $0.4844_{\pm 0.05}$ & 0.4943 & 0.5091 \\

\midrule
\midrule
SHIFT-VRM   &  & $\bm{0.5370_{\pm 0.03}}$ & \textbf{0.5697} & \textbf{0.5712} \\
\bottomrule

\end{tabular}
\end{table}


\subsection{SHIFT Allows Model Inference on Cohorts with Incomplete Data}
Tables~\ref{tab:lusc-us-full} and~\ref{tab:lusc-us-22} evaluate robustness on the US cohort for LUSC, where 175 of 197 genomic features (88.8\%) are structurally absent.
Table~\ref{tab:lusc-us-full} considers the full 197-feature setting. Here, baseline models are trained on the full TCGA feature set and rely on KNN or mean imputation at test time in the US cohort, whereas SHIFT is evaluated natively without any test-time imputation. We find that performance is highly sensitive to the imputation strategy. Under KNN imputation, most baselines degrade substantially; for example, RSF falls to an Ens-All of 0.437 and SNN to 0.422. Mean imputation partially recovers performance, with CoxPH achieving the best baseline Ens-All (0.570), followed by RSF (0.553) and DeepSurv (0.551). Within this setting, SHIFT-VRM achieves the best overall ensemble performance on US, with an Ens-All of 0.570 and a Top-3 score of 0.571, without needing any test-time imputation. These results indicate that variable-rate masking improves robustness under severe structural missingness and that masking indeed provides a practical alternative to imputation-based deployment.

Table~\ref{tab:lusc-us-22} provides a shared-feature control experiment in which all methods are trained and evaluated using only the 22 genomic features present in both TCGA and US, with no imputation. This mimics a standard approach to handle heterogeneous datasets, reducing the feature set (and potentially discarding informative features) to the lowest common denominator. In this reduced setting, SNN achieves the strongest performance (Ens-All 0.528), while SHIFT w/o VRM reaches 0.494. SHIFT-VRM, however, which is trained on the full 197 feature set and can be natively applied to the US cohort, strongly outperforms all baselines. This suggests that the benefit of SHIFT appears in the full-feature setting, where the model can exploit richer genomic inputs during training while remaining robust to severe feature absence at deployment.

\subsection{SHIFT Unlocks Multi-Site Data for Training}
\label{sec:tcgaus}

All main experiments above train exclusively on TCGA patients and evaluate generalization on external cohorts. We next consider a setting in which an additional, but incomplete, cohort is available during model development. Specifically, we test whether incorporating the 102 US patients, which contain only 22 of the 197 genomic features, improves performance in the CPTAC cohort of LUSC patients.

Table~\ref{tab:tcgaus} compares the TCGA-only SHIFT-VRM model against the combined TCGA+US SHIFT-VRM, where the US cohort is included during both training and validation. Including the incomplete US cohort during development improves overall ensemble performance on CPTAC. Compared with SHIFT-VRM (TCGA), which achieves an Ens-All C-Index of 0.602, SHIFT-VRM (TCGA+US) reaches an Ens-All of 0.629, while also yielding a validation C-index of 0.518$\pm$0.03. This result suggests that even a severely incomplete external cohort can provide useful signals during model development when its missingness is handled natively by SHIFT. More broadly, it indicates that incomplete multi-center data need not be discarded solely because large subsets of features are structurally absent.

\begin{table}[t]
\centering

\caption{\textbf{Model performance on the heterogeneous LUSC data.}
We compare SHIFT-VRM trained on TCGA (197 features) with a model trained on a heterogeneous dataset combining TCGA (197 features) and a US cohort (22 features). Performance is evaluated on the external CPTAC cohort (197 features), highlighting the benefit of model training on larger, although incomplete, cohorts.}
\label{tab:tcgaus}

\begin{tabular}{lc|ccc}
\toprule
\textbf{Method} & \textbf{Val} & \multicolumn{3}{c}{\textbf{CPTAC}} \\
\cmidrule(lr){3-5}
& Mean & Mean & Ens-All & Top-3 \\
\midrule
SHIFT-VRM (TCGA)    & $0.5170_{\pm 0.06}$ & $0.5600_{\pm 0.03}$ & 0.6021 & \textbf{0.5674} \\
SHIFT-VRM (TCGA+US) & $\bm{0.5179_{\pm 0.03}}$ & $\bm{0.5610_{\pm 0.06}}$ & \textbf{0.6287} & 0.5659 \\
\bottomrule
\end{tabular}
\end{table}

\begin{table*}[t]
\centering
\caption{\textbf{Ablation on masking fraction $f$ for SHIFT-VRM.} Validation C-index denotes mean $\pm$ std over the 5 TCGA validation folds. Bold indicates the best result in each column within each cancer block.} 
\label{tab:ablation-frac}
\resizebox{\linewidth}{!}{
\begin{tabular}{llc|ccc|ccc}
\toprule
\textbf{Method} & $\bm{f}$ & \textbf{TCGA-Val} &
\textbf{Mean} & \textbf{Ens-All} & \textbf{Top-3} &
\textbf{Mean} & \textbf{Ens-All} & \textbf{Top-3} \\
\midrule
\multicolumn{3}{c|}{} & \multicolumn{6}{c}{\textbf{GBM}} \\
\cmidrule(lr){4-9}
\multicolumn{3}{c|}{} &
\multicolumn{3}{c|}{\textbf{CPTAC}} &
\multicolumn{3}{c}{\textbf{German}} \\
\midrule
\multirow{5}{*}{SHIFT-VRM}
& 0.00 & $0.5493_{\pm 0.04}$ & $0.5390_{\pm 0.02}$ & 0.5483 & 0.5441 & $0.4926_{\pm 0.04}$ & 0.5007 & 0.4755 \\
& 0.25 & $\bm{0.5712_{\pm 0.03}}$ & $0.5644_{\pm 0.01}$ & 0.5667 & 0.5619 & $0.5247_{\pm 0.03}$ & 0.5603 & \textbf{0.5528} \\
& 0.50 & $0.5327_{\pm 0.02}$ & $0.5530_{\pm 0.01}$ & 0.5827 & \textbf{0.5839} & $0.5309_{\pm 0.01}$ & 0.5673 & 0.5470 \\
& 0.75 & $0.5214_{\pm 0.03}$ & $\bm{0.5661_{\pm 0.03}}$ & \textbf{0.5924} & \textbf{0.5839} & $0.5381_{\pm 0.02}$ & 0.5594 & 0.5430 \\
& 1.00 & $0.5297_{\pm 0.03}$ & $0.5444_{\pm 0.03}$ & 0.5640 & 0.5411 & $\bm{0.5559_{\pm 0.02}}$ & \textbf{0.5704} & 0.5510 \\
\midrule
\multicolumn{3}{c|}{} & \multicolumn{6}{c}{\textbf{LUSC}} \\
\cmidrule(lr){4-9}
\multicolumn{3}{c|}{} &
\multicolumn{3}{c|}{\textbf{CPTAC}} &
\multicolumn{3}{c}{\textbf{US (No Imputation)}} \\
\midrule
\multirow{5}{*}{SHIFT-VRM}
& 0.00 & $0.4883_{\pm 0.02}$ & $0.5597_{\pm 0.03}$ & 0.5965 & 0.5598 & $0.5102_{\pm 0.04}$ & 0.5548 & \textbf{0.5737} \\
& 0.25 & $\bm{0.5177_{\pm 0.03}}$ & $0.5535_{\pm 0.05}$ & 0.5725 & 0.5388 & $0.5347_{\pm 0.02}$ & 0.5585 & 0.5732 \\
& 0.50 & $0.5170_{\pm 0.06}$ & $\bm{0.5600_{\pm 0.03}}$ & \textbf{0.6021} & \textbf{0.5674} & $\bm{0.5370_{\pm 0.03}}$ & \textbf{0.5697} & 0.5712 \\
& 0.75 & $0.4905_{\pm 0.04}$ & $0.5224_{\pm 0.03}$ & 0.5123 & 0.5225 & $0.5272_{\pm 0.02}$ & 0.5419 & 0.5315 \\
& 1.00 & $0.5079_{\pm 0.03}$ & $0.5131_{\pm 0.03}$ & 0.5444 & 0.5444 & $0.5131_{\pm 0.01}$ & 0.5364 & 0.5362 \\
\bottomrule
\end{tabular}}
\end{table*}
\subsection[Ablation Over the Masking Fraction f]{Ablation Over the Masking Fraction $f$}
\label{sec:ablation-f}

Table~\ref{tab:ablation-frac} evaluates the effect of the masking fraction $f$ for SHIFT-VRM. Here, $f=0.0$ corresponds to no masking (i.e., SHIFT w/o VRM) and $f=1.0$ allows all features to be masked for a given training sample. We observe that for both GBM and LUSC, removing VRM tends to weaken generalization, indicating that masked training improves performance even when all features are observed. Increasing $f$ improves external robustness, though the optimal value differs across cohorts. However, $f=0.5$ provides the most consistent overall behavior and is the strongest plain SHIFT-VRM setting on the more challenging LUSC benchmark, achieving the best Ens-All on both CPTAC and US.

\section{Discussion and Conclusion}

Across two cancer types and multiple external cohorts, our results show that SHIFT is a practical approach to survival prediction from heterogeneous genomic panels without test-time imputation. The central finding is that explicitly modeling feature availability during training enables a single model to remain competitive both when feature sets are aligned and when deployed under severe structural missingness. This is particularly relevant for multi-center precision oncology, where differences in sequencing panels are common.

\textbf{Feature masking during training is competitive with imputation under severe structural missingness.} In LUSC, the US cohort lacks 175 of 197 features (88.8\%) at test time. In this setting, SHIFT-VRM essentially matches the strongest imputed baseline, CoxPH with mean imputation, and outperforms most other KNN- and mean-imputed baselines. These results suggest that, under severe panel mismatch, conditioning predictions only on observed inputs can be at least as robust as reconstructing large blocks of unmeasured features. This distinction matters because imputing a few sporadically missing values is fundamentally different from imputing features never measured by the assay. The advantage of SHIFT arises from preserving richer training information while remaining deployable on narrower panels. In the 22-feature LUSC control, SHIFT no longer outperforms the strongest baseline, suggesting that its benefit is not simply due to the transformer architecture on a small common feature set. Rather, the advantage appears in the full 197-feature setting, where SHIFT can use richer genomic information during training while remaining robust when only a limited panel is available at inference, providing a strong advantage over established baselines.

\textbf{Variable-rate masking improves external robustness beyond handling literal missingness.} The GBM experiments isolate this effect because all cohorts share the same 36 features, so no structural missingness is present at test time. In this setting, SHIFT-VRM achieves the best Ens-All on both external cohorts, whereas SHIFT w/o VRM attains the highest internal validation C-index but transfers substantially worse. This pattern suggests that VRM acts not only as a missingness simulation but also as a structured regularizer that reduces reliance on any fixed feature subset and improves cross-cohort generalization.

\textbf{Incomplete cohorts can still contribute during development.} When the incomplete US cohort of LUSC is incorporated during training and validation, SHIFT-VRM improves external CPTAC performance from an Ens-All of 0.6021 to 0.6287. This suggests that partially observed cohorts need not be excluded solely because substantial feature subsets are structurally absent. In multi-center studies, this is practically important because excluding such cohorts can reduce sample size and limit institutional diversity during model development.

\textbf{This study has several limitations.} The evaluation is restricted to two cancer types, and broader validation across additional diseases and sequencing panels is needed to establish generality. The external cohorts are modest in size, especially in the most challenging settings. We consider genomic inputs only and do not evaluate whether the same framework extends to multimodal survival prediction. In addition, the masking strategy is random rather than explicitly site-structured, whereas real panel mismatch may follow fixed institutional patterns. Finally, we focus on discrimination through the C-index and do not perform comprehensive statistical testing across all pairwise comparisons.

\textbf{Conclusion.} Overall, our findings support SHIFT as a missingness-aware survival modeling framework for heterogeneous genomic data. Across external cohorts of GBM and LUSC, SHIFT-VRM provides a strong balance of performance across both fully aligned and severely incomplete panel settings while using a single model and avoiding test-time imputation. More broadly, the results suggest that explicitly modeling structural feature missingness can support more deployable and inclusive multi-center survival prediction in precision oncology.

\bibliography{shift}

\begin{thebibliography}{41}
\providecommand{\natexlab}[1]{#1}
\providecommand{\url}[1]{\texttt{#1}}
\expandafter\ifx\csname urlstyle\endcsname\relax
  \providecommand{\doi}[1]{doi: #1}\else
  \providecommand{\doi}{doi: \begingroup \urlstyle{rm}\Url}\fi

\bibitem[Alwateer et~al.(2024)Alwateer, Atlam, Abd El-Raouf, Ghoneim, and
  Gad]{alwateer2024missing}
Majed Alwateer, El-Sayed Atlam, Mahmoud~Mohammed Abd El-Raouf, Osama~A Ghoneim,
  and Ibrahim Gad.
\newblock Missing data imputation: A comprehensive.
\newblock \emph{Journal of Computer and Communications}, 12:\penalty0 53--75,
  2024.

\bibitem[Beretta and Santaniello(2016)]{beretta2016nearest}
Lorenzo Beretta and Alessandro Santaniello.
\newblock Nearest neighbor imputation algorithms: a critical evaluation.
\newblock \emph{BMC medical informatics and decision making}, 16:\penalty0
  197--208, 2016.

\bibitem[Burgette and Reiter(2010)]{burgette2010multiple}
Lane~F Burgette and Jerome~P Reiter.
\newblock Multiple imputation for missing data via sequential regression trees.
\newblock \emph{American journal of epidemiology}, 172\penalty0 (9):\penalty0
  1070--1076, 2010.

\bibitem[Caruso et~al.(2026)Caruso, Soda, and Guarrasi]{caruso2024not}
Camillo~Maria Caruso, Paolo Soda, and Valerio Guarrasi.
\newblock Not another imputation method: A transformer-based model for missing
  values in tabular datasets.
\newblock \emph{AI Open}, 7:\penalty0 96--122, 2026.
\newblock \doi{10.1016/j.aiopen.2026.01.002}.

\bibitem[Chaudhary et~al.(2018)Chaudhary, Poirion, Lu, and
  Garmire]{chaudhary2018deep}
Kumardeep Chaudhary, Olivier~B Poirion, Liangqun Lu, and Lana~X Garmire.
\newblock Deep learning--based multi-omics integration robustly predicts
  survival in liver cancer.
\newblock \emph{Clinical cancer research}, 24\penalty0 (6):\penalty0
  1248--1259, 2018.

\bibitem[Chen et~al.(2020)Chen, Lu, Wang, Williamson, Rodig, Lindeman, and
  Mahmood]{chen2020pathomic}
Richard~J Chen, Ming~Y Lu, Jingwen Wang, Drew~FK Williamson, Scott~J Rodig,
  Neal~I Lindeman, and Faisal Mahmood.
\newblock Pathomic fusion: an integrated framework for fusing histopathology
  and genomic features for cancer diagnosis and prognosis.
\newblock \emph{IEEE Transactions on Medical Imaging}, 41\penalty0
  (4):\penalty0 757--770, 2020.

\bibitem[Chen et~al.(2021{\natexlab{a}})Chen, Lu, Chen, Williamson, and
  Mahmood]{chen2021synthetic}
Richard~J Chen, Ming~Y Lu, Tiffany~Y Chen, Drew~FK Williamson, and Faisal
  Mahmood.
\newblock Synthetic data in machine learning for medicine and healthcare.
\newblock \emph{Nature Biomedical Engineering}, 5\penalty0 (6):\penalty0
  493--497, 2021{\natexlab{a}}.

\bibitem[Chen et~al.(2021{\natexlab{b}})Chen, Lu, Weng, Chen, Williamson, Manz,
  Shady, and Mahmood]{chen2021multimodal}
Richard~J Chen, Ming~Y Lu, Wei-Hung Weng, Tiffany~Y Chen, Drew~FK Williamson,
  Trevor Manz, Maha Shady, and Faisal Mahmood.
\newblock Multimodal co-attention transformer for survival prediction in
  gigapixel whole slide images.
\newblock In \emph{Proceedings of the IEEE/CVF international conference on
  computer vision}, pages 4015--4025, 2021{\natexlab{b}}.

\bibitem[Chen et~al.(2022)Chen, Lu, Williamson, Chen, Lipkova, Noor, Shaban,
  Shady, Williams, Joo, et~al.]{chen2022pan}
Richard~J Chen, Ming~Y Lu, Drew~FK Williamson, Tiffany~Y Chen, Jana Lipkova,
  Zahra Noor, Muhammad Shaban, Maha Shady, Mane Williams, Bumjin Joo, et~al.
\newblock Pan-cancer integrative histology-genomic analysis via multimodal deep
  learning.
\newblock \emph{Cancer Cell}, 40\penalty0 (8):\penalty0 865--878, 2022.

\bibitem[Chen et~al.(2023)Chen, Wang, Williamson, Chen, Lipkova, Lu, Sahai, and
  Mahmood]{chen2023algorithmic}
Richard~J Chen, Judy~J Wang, Drew~FK Williamson, Tiffany~Y Chen, Jana Lipkova,
  Ming~Y Lu, Sharifa Sahai, and Faisal Mahmood.
\newblock Algorithmic fairness in artificial intelligence for medicine and
  healthcare.
\newblock \emph{Nature biomedical engineering}, 7\penalty0 (6):\penalty0
  719--742, 2023.

\bibitem[Chen and Guestrin(2016)]{10.1145/2939672.2939785}
Tianqi Chen and Carlos Guestrin.
\newblock Xgboost: A scalable tree boosting system.
\newblock In \emph{Proceedings of the 22nd ACM SIGKDD International Conference
  on Knowledge Discovery and Data Mining}, KDD '16, page 785–794, New York,
  NY, USA, 2016. Association for Computing Machinery.
\newblock ISBN 9781450342322.
\newblock \doi{10.1145/2939672.2939785}.

\bibitem[Chen et~al.(2024)Chen, Shen, Feng, and Panageas]{chen2024unlocking}
Yuan Chen, Ronglai Shen, Xiwen Feng, and Katherine Panageas.
\newblock Unlocking the power of multi-institutional data: Integrating and
  harmonizing genomic data across institutions.
\newblock \emph{Biometrics}, 80\penalty0 (4):\penalty0 ujae146, 2024.

\bibitem[Cohen et~al.(2018)Cohen, Luck, and Honari]{cohen2018distribution}
Joseph~Paul Cohen, Margaux Luck, and Sina Honari.
\newblock Distribution matching losses can hallucinate features in medical
  image translation.
\newblock In \emph{Medical Image Computing and Computer Assisted
  Intervention--MICCAI 2018: 21st International Conference, Granada, Spain,
  September 16-20, 2018, Proceedings, Part I}, pages 529--536. Springer, 2018.

\bibitem[Cox(1972)]{cox1972regression}
David~R Cox.
\newblock Regression models and life-tables.
\newblock \emph{Journal of the Royal Statistical Society: Series B
  (Methodological)}, 34\penalty0 (2):\penalty0 187--202, 1972.

\bibitem[Donders et~al.(2006)Donders, Van Der~Heijden, Stijnen, and
  Moons]{donders2006gentle}
A~Rogier~T Donders, Geert~JMG Van Der~Heijden, Theo Stijnen, and Karel~GM
  Moons.
\newblock A gentle introduction to imputation of missing values.
\newblock \emph{Journal of clinical epidemiology}, 59\penalty0 (10):\penalty0
  1087--1091, 2006.

\bibitem[Edwards et~al.(2015)Edwards, Oberti, Thangudu, Cai, McGarvey, Jacob,
  Madhavan, and Ketchum]{edwards2015cptac}
Nathan~J Edwards, Mauricio Oberti, Ratna~R Thangudu, Shuang Cai, Peter~B
  McGarvey, Shine Jacob, Subha Madhavan, and Karen~A Ketchum.
\newblock The cptac data portal: a resource for cancer proteomics research.
\newblock \emph{Journal of proteome research}, 14\penalty0 (6):\penalty0
  2707--2713, 2015.

\bibitem[Fan et~al.(2024)Fan, Feng, Wu, Liu, and Jiang]{fan2024multiscale}
Yuwei Fan, Chenlong Feng, Rui Wu, Chao Liu, and Dongxiang Jiang.
\newblock Multiscale-attention masked autoencoder for missing data imputation
  of wind turbines.
\newblock \emph{Knowledge-Based Systems}, 299:\penalty0 112114, 2024.

\bibitem[Flores et~al.(2023)Flores, Claborne, Weller, Webb-Robertson, Waters,
  and Bramer]{flores2023missing}
Javier~E Flores, Daniel~M Claborne, Zachary~D Weller, Bobbie-Jo~M
  Webb-Robertson, Katrina~M Waters, and Lisa~M Bramer.
\newblock Missing data in multi-omics integration: Recent advances through
  artificial intelligence.
\newblock \emph{Frontiers in artificial intelligence}, 6:\penalty0 1098308,
  2023.

\bibitem[Forestier et~al.(2017)Forestier, Petitjean, Dau, Webb, and
  Keogh]{forestier2017generating}
Germain Forestier, Fran{\c{c}}ois Petitjean, Hoang~Anh Dau, Geoffrey~I Webb,
  and Eamonn Keogh.
\newblock Generating synthetic time series to augment sparse datasets.
\newblock In \emph{2017 IEEE international conference on data mining (ICDM)},
  pages 865--870. IEEE, 2017.

\bibitem[Garcia et~al.(2017)Garcia, Minkovsky, Jia, Ducar, Shivdasani, Gong,
  Ligon, Sholl, Kuo, MacConaill, et~al.]{garcia2017validation}
Elizabeth~P Garcia, Alissa Minkovsky, Yonghui Jia, Matthew~D Ducar, Priyanka
  Shivdasani, Xin Gong, Azra~H Ligon, Lynette~M Sholl, Frank~C Kuo, Laura~E
  MacConaill, et~al.
\newblock Validation of oncopanel: a targeted next-generation sequencing assay
  for the detection of somatic variants in cancer.
\newblock \emph{Archives of Pathology and Laboratory Medicine}, 141\penalty0
  (6):\penalty0 751--758, 2017.

\bibitem[Heymans and Twisk(2022)]{heymans2022handling}
Martijn~W Heymans and Jos~WR Twisk.
\newblock Handling missing data in clinical research.
\newblock \emph{Journal of clinical epidemiology}, 151:\penalty0 185--188,
  2022.

\bibitem[Ishwaran et~al.(2008)Ishwaran, Kogalur, Blackstone, and
  Lauer]{Ishwaran_2008}
Hemant Ishwaran, Udaya~B. Kogalur, Eugene~H. Blackstone, and Michael~S. Lauer.
\newblock {Random survival forests}.
\newblock \emph{The Annals of Applied Statistics}, 2\penalty0 (3):\penalty0 841
  -- 860, 2008.
\newblock \doi{10.1214/08-AOAS169}.

\bibitem[Jaume et~al.(2024)Jaume, Vaidya, Chen, Williamson, Liang, and
  Mahmood]{jaume2024modeling}
Guillaume Jaume, Anurag Vaidya, Richard~J Chen, Drew~FK Williamson, Paul~Pu
  Liang, and Faisal Mahmood.
\newblock Modeling dense multimodal interactions between biological pathways
  and histology for survival prediction.
\newblock In \emph{Proceedings of the IEEE/CVF Conference on Computer Vision
  and Pattern Recognition}, pages 11579--11590, 2024.

\bibitem[Joel et~al.(2022)Joel, Doorsamy, and Paul]{joel2022review}
Luke~Oluwaseye Joel, Wesley Doorsamy, and Babu~Sena Paul.
\newblock A review of missing data handling techniques for machine learning.
\newblock \emph{International Journal of Innovative Technology and
  Interdisciplinary Sciences}, 5\penalty0 (3):\penalty0 971--1005, 2022.

\bibitem[Katzman et~al.(2018)Katzman, Shaham, Cloninger, Bates, Jiang, and
  Kluger]{katzman2018deepsurv}
Jared~L Katzman, Uri Shaham, Alexander Cloninger, Jonathan Bates, Tingting
  Jiang, and Yuval Kluger.
\newblock Deepsurv: personalized treatment recommender system using a cox
  proportional hazards deep neural network.
\newblock \emph{BMC medical research methodology}, 18:\penalty0 1--12, 2018.

\bibitem[Klambauer et~al.(2017)Klambauer, Unterthiner, Mayr, and
  Hochreiter]{klambauer2017self}
G{\"u}nter Klambauer, Thomas Unterthiner, Andreas Mayr, and Sepp Hochreiter.
\newblock Self-normalizing neural networks.
\newblock In \emph{Proceedings of the 31st international conference on neural
  information processing systems}, pages 972--981, 2017.

\bibitem[Lee et~al.(2018)Lee, Zame, Yoon, and Van Der~Schaar]{lee2018deephit}
Changhee Lee, William Zame, Jinsung Yoon, and Mihaela Van Der~Schaar.
\newblock Deephit: A deep learning approach to survival analysis with competing
  risks.
\newblock In \emph{Proceedings of the AAAI conference on artificial
  intelligence}, volume~32, 2018.

\bibitem[Lee(2023)]{lee2023deep}
Minhyeok Lee.
\newblock Deep learning techniques with genomic data in cancer prognosis: a
  comprehensive review of the 2021--2023 literature.
\newblock \emph{Biology}, 12\penalty0 (7):\penalty0 893, 2023.

\bibitem[Lipkova et~al.(2022)Lipkova, Chen, Chen, Lu, Barbieri, Shao, Vaidya,
  Chen, Zhuang, Williamson, et~al.]{lipkova2022artificial}
Jana Lipkova, Richard~J Chen, Bowen Chen, Ming~Y Lu, Matteo Barbieri, Daniel
  Shao, Anurag~J Vaidya, Chengkuan Chen, Luoting Zhuang, Drew~FK Williamson,
  et~al.
\newblock Artificial intelligence for multimodal data integration in oncology.
\newblock \emph{Cancer cell}, 40\penalty0 (10):\penalty0 1095--1110, 2022.

\bibitem[Neog et~al.(2025)Neog, Daw, Khorasgani, and Karpatne]{neog2025masking}
Abhilash Neog, Arka Daw, Sepideh~Fatemi Khorasgani, and Anuj Karpatne.
\newblock Masking the gaps: An imputation-free approach to time series modeling
  with missing data.
\newblock \emph{arXiv preprint arXiv:2502.15785}, 2025.

\bibitem[Petrazzini et~al.(2021)Petrazzini, Naya, Lopez-Bello, Vazquez, and
  Spangenberg]{petrazzini2021evaluation}
Ben~Omega Petrazzini, Hugo Naya, Fernando Lopez-Bello, Gustavo Vazquez, and
  Luc{\'\i}a Spangenberg.
\newblock Evaluation of different approaches for missing data imputation on
  features associated to genomic data.
\newblock \emph{BioData mining}, 14:\penalty0 1--13, 2021.

\bibitem[Pujianto et~al.(2019)Pujianto, Wibawa, Akbar, et~al.]{pujianto2019k}
Utomo Pujianto, Aji~Prasetya Wibawa, Muhammad~Iqbal Akbar, et~al.
\newblock K-nearest neighbor (k-nn) based missing data imputation.
\newblock In \emph{2019 5th International Conference on Science in Information
  Technology (ICSITech)}, pages 83--88. IEEE, 2019.

\bibitem[Tomczak et~al.(2015)Tomczak, Czerwi{\'n}ska, and Wiznerowicz]{TCGA}
Katarzyna Tomczak, Patrycja Czerwi{\'n}ska, and Maciej Wiznerowicz.
\newblock Review the cancer genome atlas (tcga): an immeasurable source of
  knowledge.
\newblock \emph{Contemporary Oncology/Wsp{\'o}{\l}czesna Onkologia},
  2015\penalty0 (1):\penalty0 68--77, 2015.

\bibitem[Vale-Silva and Rohr(2021)]{vale2021long}
Lu{\'\i}s~A Vale-Silva and Karl Rohr.
\newblock Long-term cancer survival prediction using multimodal deep learning.
\newblock \emph{Scientific Reports}, 11\penalty0 (1):\penalty0 13505, 2021.

\bibitem[Vaswani et~al.(2017)Vaswani, Shazeer, Parmar, Uszkoreit, Jones, Gomez,
  Kaiser, and Polosukhin]{vaswani2017attention}
Ashish Vaswani, Noam Shazeer, Niki Parmar, Jakob Uszkoreit, Llion Jones,
  Aidan~N Gomez, {\L}ukasz Kaiser, and Illia Polosukhin.
\newblock Attention is all you need.
\newblock \emph{Advances in neural information processing systems}, 30, 2017.

\bibitem[Wiegrebe et~al.(2024)Wiegrebe, Kopper, Sonabend, Bischl, and
  Bender]{wiegrebe2024deep}
Simon Wiegrebe, Philipp Kopper, Raphael Sonabend, Bernd Bischl, and Andreas
  Bender.
\newblock Deep learning for survival analysis: a review.
\newblock \emph{Artificial Intelligence Review}, 57\penalty0 (3):\penalty0 65,
  2024.

\bibitem[Yi et~al.(2019)Yi, Walia, and Babyn]{yi2019generative}
Xin Yi, Ekta Walia, and Paul Babyn.
\newblock Generative adversarial network in medical imaging: A review.
\newblock \emph{Medical image analysis}, 58:\penalty0 101552, 2019.

\bibitem[Yousefi et~al.(2017)Yousefi, Amrollahi, Amgad, Dong, Lewis, Song,
  Gutman, Halani, Velazquez~Vega, Brat, et~al.]{yousefi2017predicting}
Safoora Yousefi, Fatemeh Amrollahi, Mohamed Amgad, Chengliang Dong, Joshua~E
  Lewis, Congzheng Song, David~A Gutman, Sameer~H Halani, Jose~Enrique
  Velazquez~Vega, Daniel~J Brat, et~al.
\newblock Predicting clinical outcomes from large scale cancer genomic profiles
  with deep survival models.
\newblock \emph{Scientific reports}, 7\penalty0 (1):\penalty0 1--11, 2017.

\bibitem[Zadeh and Schmid(2020)]{zadeh2020bias}
Shekoufeh~Gorgi Zadeh and Matthias Schmid.
\newblock Bias in cross-entropy-based training of deep survival networks.
\newblock \emph{IEEE transactions on pattern analysis and machine
  intelligence}, 43\penalty0 (9):\penalty0 3126--3137, 2020.

\bibitem[Zhang et~al.(2024)Zhang, Wang, Peng, and Duan]{zhang2024improved}
Zhi-cheng Zhang, Yong Wang, Jian-jian Peng, and Jun-ting Duan.
\newblock An improved self-attention for long-sequence time-series data
  forecasting with missing values.
\newblock \emph{Neural computing and applications}, 36\penalty0 (8):\penalty0
  3921--3940, 2024.

\bibitem[Zhou et~al.(2024)Zhou, Aryal, and Bouadjenek]{zhou2024review}
Youran Zhou, Sunil Aryal, and Mohamed~Reda Bouadjenek.
\newblock Review for handling missing data with special missing mechanism.
\newblock \emph{arXiv preprint arXiv:2404.04905}, 2024.

\end{thebibliography}
\end{document}